# Scale-Rotation-Equivariant Lie Group Convolution Neural Networks (Lie Group-CNNs)

Wei-Dong Qiao, Yang Xu, and Hui Li

**Abstract**—The weight-sharing mechanism of convolutional kernels ensures translation-equivariance of convolution neural networks (CNNs). Recently, rotation-equivariance has been investigated. However, research on scale-equivariance or simultaneous scale-rotation-equivariance is insufficient. This study proposes a Lie group-CNN, which can keep scale-rotation-equivariance for image classification tasks. The Lie group-CNN includes a lifting module, a series of group convolution modules, a global pooling layer, and a classification layer. The lifting module transfers the input image from Euclidean space $\mathbb{R}^n$ to Lie group space, and the group convolution is parameterized through a fully connected network using Lie-algebra of Lie-group elements as inputs to achieve scale-rotation-equivariance. The Lie group SIM(2) is utilized to establish the Lie group-CNN with scale-rotation-equivariance. Scale-rotation-equivariance of Lie group-CNN is verified and achieves the best recognition accuracy on the blood cell dataset (97.50%) and the HAM10000 dataset (77.90%) superior to Lie algebra convolution network, dilation convolution, spatial transformer network, and scale-equivariant steerable network. In addition, the generalization ability of the Lie group-CNN on SIM(2) on rotation-equivariance is verified on rotated-MNIST and rotated-CIFAR10, and the robustness of the network is verified on SO(2) and SE(2). Therefore, the Lie group-CNN can successfully extract geometric features and performs equivariant recognition on images with rotation and scale transformations.

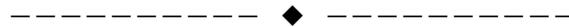

## 1 INTRODUCTION

CONVOLUUTIONAL neural networks have achieved unprecedented prosperity in the computer vision community [1] and become one of the most successful deep learning models achieving state-of-the-art progress on multiple tasks, such as image classification, object detection, and image segmentation. For example, VGG utilized small convolution filters instead of large ones and increased the network depth [2]. ResNet proposed a residual learning module to address the vanishing gradient issue in deep networks [3]. Later, DenseNet connected each layer to the others in a feed-forward method [4]. The translation equivariance is a critical performance of CNN. It is usually achieved by the weight-sharing mechanism of convolutional filters and the pooling operation to generate global features in the conventional CNN. The convolutional filters can preserve the same features before and after the translation transformations using identical weights to convolute local regions of the input image with sliding strides. Therefore, the regular CNN automatically possesses the translation symmetry. In addition, Lenc and Vedaldi [5] found that the first several layers of CNN could be predictably transformed with image warping and regarded as warping-equivariant.

Besides transformation equivariance, when the relative motion occurs between a monocular camera and an object due to translation, scale, and rotation transformations, the captured image is supposed to change accordingly. Extracting invariant features and ensuring accurate predictions from images before and after translation, scale, and rotation transformations are enduring issues in computer vision. An equivariant feature detector guarantees that the output would transform the same way as the input transforms, ensuring no feature and information leakage occur. The invariance is a particular case of equivariance, i.e., the output would remain the same no matter what changes occurred in the input. Theoretically, the input images under different transformations would lead CNN to learn the invariant features adaptively. However, the unified equivariance of CNN could only be given empirically, and previous experiments validated that the unified equivariance of CNN was fragile when the input images were translated, scaled, and rotated [6].

Many classical image recognition algorithms achieve equivariance by detecting features that are insensitive to input transformations. For example, the image pyramid was proposed by downsampling the input images using a series of Gaussian convolutions with various scale parameters, and the local features with scale and rotation invariance were extracted in the scale space [7]. Lowe [8] proposed the Scale Invariant Feature Transform (SIFT) algorithm to construct a Difference of Gaussian (DoG) pyramid and find key points that do not change with image transformations. In addition to establishing the image pyramid and feature pyramid, another way to achieve equivariance is to design steerable filters. For example, arbitrary rotation of the filter response was interpolated from linear

―――――――――
- *W.D. Qiao, Y. Xu, and H, Li are with Key Lab of Smart Prevention and Mitigation of Civil Engineering Disasters of the Ministry of Industry and Information Technology, Harbin Institute of Technology, Harbin, 150090, China.*
- *Y. Xu and H, Li are with Laboratory of Artificial Intelligence, Harbin Institute of Technology, Harbin, 150001, China.*
- *H, Li is the corresponding author (lihui@hit.edu.cn).*

combinations of basis filters [9], and the steerable pyramid decomposed the image into a series of image subbands with different scales and orientations using basis functions of Gabor wavelets [10], which was designed to be scale and rotation invariant separately. Furthermore, a wavelet scattering network with translation, scale, and rotation invariances was proposed using scaled and rotated wavelets as filters [11,12,13]. However, the equivariant or invariant filters in the above mentioned methods are hand-crafted with prior knowledge and cannot be autonomously learned.

To overcome the theoretical limitations of investigating equivariance for the regular CNN in the $\mathbb{R}^n$, the group theory provides a promising way to construct a unique form of group CNN with equivariance in the group space [14,15]. In this paper, the group convolution is designed with scale and rotation equivariance in the group space by exploiting Lie groups with scaling and rotational group elements. The main contributions of this study are summarized as follows.

(1) It is theoretically derived that the regular CNN does not possess the scale and rotation equivariance while the proposed Lie-group convolution does;

(2) Lie group of SIM(2) and the corresponding Lie algebras with scale and rotation transformation elements are introduced to construct the Lie group-CNN with scale and rotation equivariance;

(3) The scale and rotation equivariance Lie-group CNN mainly consists of a lifting module and several group convolution modules, in which the lifting module is designed to transfer the input image into the Lie group space, and the group convolution is parameterized by a fully connected network using the Lie algebra as input;

(4) Finally, the scale and rotation equivariance of the Lie group-CNN is verified on two scaled-rotated image datasets. In addition, the generalization ability of the method on rotation equivariance is also verified on two rotated image datasets and the robustness of the network is verified on SO(2) and SE(2) with rotation transformations. Comparative studies are performed to demonstrate the efficacy of the proposed method superior to existing dilation convolution, STN, SESN, L-Conv, and multilayer perceptron (MLP)-based neural networks.

The remainder of this paper is arranged as follows. Section 2 introduces some related work on the scale equivariance and rotation equivariance of CNN. Section 3 gives mathematical formulations of equivariance and theoretical proofs of lacking scale and rotation equivariance in regular CNNs. Section 4 describes the proposed scale and rotation equivariant Lie group-CNN applying Lie groups and Lie algebra, including the overall architecture and the novel modules of lifting operation, group convolution, and parameterization strategy. Section 5 shows the experimental results and comparative studies on two scaled-rotated and two rotated image datasets. Section 6 concludes the paper.

## 2 RELATED WORK

The equivariance of CNN has been investigated under various forms of translation, rotation, and scale transformations. Anti-aliasing and downsampling strategies were integrated to improve the shift equivariance of CNN [16]. The rotation equivariance of CNN is gained mainly by introducing rotated features of the input images. For example, a specialized CNN was designed to autonomously learn abstract representations of the images and obtain the rotational symmetry in galaxy images [17]. A rotation-invariance layer was introduced to guide the training process to learn similar features before and after rotation [18]. RotConv was proposed by rotating convolutional kernels with multiple orientations, and a vector field representing magnitude and angle at every spatial location was generated [19]. A gradient-aligned convolution was proposed to perform a pixel-level gradient alignment operation before the regular convolution and achieve rotation equivariance [20]. Xie et al. proposed a filter parametrization method based on an ameliorated Fourier series expansion for 2D filters to construct rotation equivariant convolutions [21]. As for the scale equivariance, the multi-scale feature pyramid and dilated convolution were utilized to establish the scale-invariance CNN [22,23] and they were also used in the field of object detection and image segmentation for multi-scale object [24,25] Gao et al. [26] proposed Res2Net representing multi-scale features at a granular level by constructing hierarchical residual-like connections in a single residual block. Jaderberg et al. [27] introduced a spatial transformation network where a spatial transformer module was designed to produce an appropriate spatial transformation for each input image, and the transformations were performed on all feature maps to achieve spatial equivariance. Qi et al. [28] presented deterministic autoencoding transformation and probabilistic autoencoding variational transformation models to learn more generic visual representations.

Since transformations of input images can be regarded as group elements, the group theory has been introduced into CNNs to extend various equivariance properties, and most previous work mainly focuses on building equivariant architectures using discrete groups. The group convolution was defined through a surjective exponential mapping with the Lie algebra on the Lie group [15]. The equivariance of the neural network layer could be achieved by parameter-sharing under any operation of discrete groups [29]. Standard convolution kernels were transformed by discrete groups of 90-degree rotations and reflections [30,31]. General principles of constructing convolutions in a compact group and integrating over the group space were introduced [32]. Bekkers [33] proposed that using a generic basis of B-splines defined by the Lie algebra could express the equivariance of convolution kernels, and irreducible group representations were found to bring constraints on the expression ability of convolutional kernels [34]. Sosnovik et al. [35] developed a scale-equivariant steerable network using the steerable filter parametrization to scale the filter instead of resizing the tensor. Dehmamy et al. [36] proposed the Lie algebra convolutional network that can automatically discover symmetries with Lie algebra instead of Lie groups.

## 3 MATHEMATICAL FORMULATION OF EQUIVARIANCE

The equivariance of a mapping function $\Phi(\cdot)$ on an input image $I$ is defined as

$$\Phi(T_g I) = T_g \Phi(I) \tag{1}$$

where $T_g$ denotes a linear transformation matrix corresponding to a specific transformation operation $g$. $g \in G$, where $G$ is a group consisting of all the possible group elements. $I \in \mathbb{R}^n$, $\Phi : \mathbb{R}^n \to \mathbb{R}^m$, and $T_g$ acts on the input and output spaces. If (1) holds to $\forall I \in \mathbb{R}^n, \forall g \in G$, $\Phi$ is equivariant to $G$. According to the linear transformation, for two different transformation operations $g$ and $g'$, $T_{gg'} = T_g T_{g'}$. Note that the equivariance makes the mapping function $\Phi$ symmetrical to a certain transformation operation $g$, thus improving the generalization ability and geometric reasoning ability under various transformations. By the way, invariance is a special case of equivariance, i.e., $\Phi(T_g I) = \Phi(I)$, where the output remains precisely the same no matter what transformation has been applied to the input.

Specifically, for the scale and rotation equivariance, $m = (s, \theta)$ can be expressed as a combination of a random scale factor $s$ and a random rotation angle $\theta$, where $s > 0$ and $\theta \in \mathbb{R}$. The matrix form acting on the image coordinate system is $A_m = \begin{bmatrix} a & b \\ c & d \end{bmatrix}$ and $A_m$ can be represented as the product of a rotation matrix $R_\theta$ and an isotropic scale matrix $Z_s$

$$A_m = \begin{bmatrix} a & b \\ c & d \end{bmatrix} = \begin{bmatrix} s \cdot \cos\theta & -s \cdot \sin\theta \\ s \cdot \sin\theta & s \cdot \cos\theta \end{bmatrix} = \begin{bmatrix} s & 0 \\ 0 & s \end{bmatrix} \cdot \begin{bmatrix} \cos\theta & -\sin\theta \\ \sin\theta & \cos\theta \end{bmatrix} = Z_s R_\theta \tag{2}$$

When a two-dimensional (2D) image is scaled-rotated-transformed around its center, the transformed image remains the same pixel intensity as the original image:

$$I'(x') = I(x) \tag{3}$$

where $I'(x')$ denotes the intensity values of the transformed image in the transformed coordinates $(x')$, and $I(x)$ denotes the intensity values of the original image in the original coordinates $(x)$. The coordinates before and after the scaled and rotated transformation satisfy $x' = A_m x$. (3) can be written as

$$I'(x) = I(A_m^{-1} x) \tag{4}$$

Next, the mathematical formula is derived to demonstrate that CNN is not scale and rotation equivariance. For an input layer or a hidden layer, a regular convolution operator can be calculated as

$$[f * \psi](x) = \sum_{\tilde{y}} f(x - \tilde{y}) \psi(\tilde{y}) = \sum_y f(y) \psi(x - y) \tag{5}$$

where $f$ denotes the feature map from the last layer, $\psi$ denotes the convolution kernel, $x$ denotes the input coordinates, $\tilde{y}$ denotes the coordinates of the convolution kernel.

Considering that the feature map is scaled-rotated-transformed, it can be derived by (4):

$$T_m f(x) = f(A_m^{-1} x) \tag{6}$$

The scale and rotation equivariance can be derived by

$$[(T_m f) * \psi](x) = \sum_y f(A_m^{-1} y) \psi(x - y) = \sum_y f(y) \psi(x - A_m y) = \sum_y f(y) \psi[A_m(A_m^{-1} x - y)] \tag{7}$$

Based on (5) and (6), the following equation can be derived as

$$T_m[f * (T_{m^{-1}} \psi)](x) = [f * (T_{m^{-1}} \psi)](A_m^{-1} x) = \sum_y f(y)[(T_{m^{-1}} \psi)(A_m^{-1} x - y)] = \sum_y f(y) \psi[A_m(A_m^{-1} x - y)] \tag{8}$$

where $T_{m^{-1}}$ denotes the inverse scaled-rotated-transformed.

Note that the right-side terms of (7) and (8) are the same. Then we have

$$[(T_m f) * \psi](x) = T_m[f * (T_{m^{-1}} \psi)](x) \tag{9}$$

Equation (9) indicates that CNN is not scale and rotation equivariance because there is an additional operator of $T_{m^{-1}}$ acting on the convolution kernel $\psi$ on the right side, as the convolution of a scaled and rotated feature map is equal to the scale and rotation of an inverse scaled and rotated convolution.

After the convolution operation, the feature maps are inputted into the nonlinear activation (e.g., ReLU), batch normalization (BN), linear, and global average pooling (GAP) layers:

ReLU: $\quad ReLU(f_i) = \max(0, f_i)$

BN: $\quad \mu_B = \frac{1}{\mathcal{M}} \sum_{i=1}^{\mathcal{M}} f_i, \sigma_B^2 = \frac{1}{\mathcal{M}} \sum_{i=1}^{\mathcal{M}} (f_i - \mu_B)^2,$

$$\hat{f}_i = \frac{f_i - \mu_B}{\sqrt{\sigma_B^2 + \epsilon}}, f_i' = \gamma \hat{f}_i + \beta, f' = W * f + b \tag{10}$$

Linear: $\quad f' = W * f + b$

GAP: $\quad f'' = \frac{1}{\mathcal{N}} \sum_i f_i$

where $f_i$ denotes feature value at position $i$; $\mu_B$ and $\sigma_B^2$ denote mean and variance; $\mathcal{M}$ denotes batch size; $\gamma$ and $\beta$ are hyperparameters; $\hat{f}_i$ and $f_i'$ denote the results after normalization and after scaling and translation, respectively; $W$ and $b$ denote weights and bias; $f'$ denotes the result after the linear layer; $\mathcal{N}$ and $f''$ denote the number and the mean value of pixels in feature map, respectively.

As shown in (10), the ReLU, BN, linear, and global average pooling layers are scale and rotation equivariance because the inputs are values after convolution.

# 4 METHODOLOGY

In this section, a novel Lie group-CNN is proposed based on the scale and rotation equivariance group convolution. Firstly, the Lie group and Lie algebra are introduced to solve the equivariant issue of regular CNNs. Then, the lifting layer is designed for mapping images to Lie algebras from the $\mathbb{R}^n$ to the Lie group space. Finally, the scale and rotation equivariance group convolution in the Lie group space is derived and parameterization as a fully connected network.

## 4.1 Lie Groups and Lie Algebras

Many sets of transformations satisfy the group properties and thus can form a group. Let $G$ denote a group under a single operation with the following properties:

1) For any three elements $g_1, g_2, g_3 \in G$, they satisfy the associativity $(g_1 g_2) g_3 = g_1 (g_2 g_3)$.

2) The composition of any two elements $g_1 g_2 \in G$.

3) Each element $g_1$ has a unique inverse $g_1^{-1}$, and $g_1^{-1} \in G$.

4) There is a unique unit element $e$ in the group, i.e., $g_1 g_1^{-1} = e$.

Take the isotropic scale group in the 2D space as an example. Associativity can be verified by $(Z_{s_1} Z_{s_2}) Z_{s_3} = Z_{s_1} (Z_{s_2} Z_{s_3})$. The multiplication of two scale matrices forms the third scale matrix, i.e., $Z_{s_1} Z_{s_2} = Z_{s_1 s_2}$. The inverse of a scale matrix $Z_s$ is $Z_{1/s}$. The unit element of identity is the scale factor $s = 1$.

The Lie group $G$ is the group with continuous, differentiable, and smooth. The similar transformation group SIM(2) contains scale and rotation transformation and the other two Lie groups (the special orthogonal group SO(2) and special Euclidean group SE(2)) with rotation transformation, all of which can be expressed by nonsingular real matrices in the 2D space. Since the Lie group is not defined in the vector space, it is challenging to directly measure the distance between group elements. Therefore, the Lie algebra $\mathfrak{g}$ is utilized to describe the local properties in the tangent plane around the identity of $G$ by infinitesimal transformations. The Lie algebra $\mathfrak{g}$ corresponds to the $k$-dimensional vector space, i.e., $(v_1, \ldots, v_k) \in \mathfrak{g}$, with the corresponding basis matrix elements $\{e_1, \ldots, e_k\}$. An expanded matrix $\Lambda$ can be generated by

$$\Lambda = \sum_{i=1}^{k} v_i e_i \tag{11}$$

where $v_i$ is the $i$-th component of the $k$-dimensional vector $(v_1, \ldots, v_k)$.

For Lie groups, the matrix exponentiation is used to map the Lie algebra $\mathfrak{g}$ to the group $G$, i.e., $\exp(\Lambda) = \sum_{n=0}^{\infty} \frac{\Lambda^n}{n!} : \mathfrak{g} \to G$. which has a closed form and an inverse of logarithm mapping as $\log : G \to \mathfrak{g}$. Take the isotropic scale group as an example to describe the relationship between group $G$ and the Lie algebra $\mathfrak{g}$. The scale element $\lambda = (v_1, \ldots, v_k)_{k=1}$ is the Lie algebra of the group element $Z_s$, $e_1$ is the basis matrix element, and $Z_s = \exp(\lambda e_1)$ is the Lie algebra parametrization of the isotropic scale group. The multiplication of group elements can be achieved by the addition of Lie algebra, i.e., $Z_{s_1} Z_{s_2} = \exp(\lambda_1 e_1) \exp(\lambda_2 e_1) = \exp[(\lambda_1 + \lambda_2) e_1] = Z_{s_1 s_2}$. The specific form of the similar transformation group SIM(2) is shown below.

The SIM(2) group is the group of similarity transformations in the 2D space, i.e.,

$$\text{SIM}(2) = \left\{ G = \begin{bmatrix} s \cdot \cos\theta & -s \cdot \sin\theta & t_u \\ s \cdot \sin\theta & s \cdot \cos\theta & t_w \\ 0 & 0 & 1 \end{bmatrix} = \begin{bmatrix} sR^{2\times 2} & t^{2\times 1} \\ 0^{1\times 2} & 1 \end{bmatrix} \right\} \in \mathbb{R}^{3 \times 3}$$

, where $R$ represents the rotation transformation, $t \in \mathbb{R}^2$ represents the translation transformation, and $s \in \mathbb{R}^+$ denotes the scale transformation for uniformly zoom-in ($s > 1$) and zoom-out ($0 < s < 1$) operations. Using the similarity transformation of the SIM(2) group, the similarity transformation can be derived by the homogeneous coordinates as

$$\begin{pmatrix} U' \\ W' \\ 1 \end{pmatrix} = \begin{pmatrix} U(s \cdot \cos\theta) - W(s \cdot \sin\theta) + t_u \\ U(s \cdot \sin\theta) + W(s \cdot \cos\theta) + t_w \\ 1 \end{pmatrix} = \begin{bmatrix} s \cdot \cos\theta & -s \cdot \sin\theta & t_u \\ s \cdot \sin\theta & s \cdot \cos\theta & t_w \\ 0 & 0 & 1 \end{bmatrix} \begin{pmatrix} U \\ W \\ 1 \end{pmatrix} \tag{12}$$

where $t^{2\times 1} = \begin{pmatrix} t_u \\ t_w \end{pmatrix}$ denotes the translation vector, $U$ and $W$ denote the 2D coordinates in the original image, $U'$ and $W'$ denote the 2D coordinates in the transformed image.

The Lie algebra $\mathfrak{sim}(2)$ of SIM(2) is a four-dimensional vector, i.e., $\boldsymbol{v} = (u, w, \theta, \lambda) \in \mathbb{R}^4$. Therefore, $\mathfrak{sim}(2)$ has four basis matrix elements:

$$e_1 = \begin{bmatrix} 0 & 0 & 1 \\ 0 & 0 & 0 \\ 0 & 0 & 0 \end{bmatrix}, \quad e_2 = \begin{bmatrix} 0 & 0 & 0 \\ 0 & 0 & 1 \\ 0 & 0 & 0 \end{bmatrix}, \quad e_3 = \begin{bmatrix} 0 & -1 & 0 \\ 1 & 0 & 0 \\ 0 & 0 & 0 \end{bmatrix},$$

$e_4 = \begin{bmatrix} 0 & 0 & 0 \\ 0 & 0 & 0 \\ 0 & 0 & -1 \end{bmatrix}$. The matrix exponentiation mapping from $\mathfrak{sim}(2)$ to SIM(2) is expressed using the Taylor series as

$$\exp\left(\sum_{i=1}^{4} v_i e_i\right) = \sum_{n=0}^{\infty} \frac{1}{n!} \begin{bmatrix} 0 & -\theta & u \\ \theta & 0 & w \\ 0 & 0 & -\lambda \end{bmatrix}^n = \begin{bmatrix} R & B \cdot \begin{pmatrix} u \\ w \end{pmatrix} \\ 0 & e^{-\lambda} \end{bmatrix} \tag{13}$$

$B = \begin{bmatrix} C & -\theta \cdot D \\ \theta \cdot D & C \end{bmatrix},$

$$C = \left(\frac{\lambda^2}{\lambda^2+\theta^2}\right)\cdot\left(\frac{1-e^{-\lambda}}{\lambda}\right) + \left(\frac{\theta^2}{\lambda^2+\theta^2}\right)\cdot\left(\frac{sin\theta}{\theta} - \lambda\frac{1-cos\theta}{\theta^2}\right),$$

$$D = \left(\frac{\lambda^2}{\lambda^2+\theta^2}\right)\cdot\left(\frac{e^{-\lambda}-1+\lambda}{\lambda^2}\right) + \left(\frac{\theta^2}{\lambda^2+\theta^2}\right)\cdot\left(\frac{1-cos\theta}{\theta^2} - \lambda\frac{\theta-sin\theta}{\theta^3}\right)$$

Conversely, to obtain $\mathfrak{sim}(2)$ with a known SIM(2), the rotation element $\theta$ can be calculated by trigonometric function; the scale element $\lambda$ can be obtained by the negative logarithmic operation, and the translation elements $u$ and $w$ can be determined by

$$\begin{pmatrix}u\\w\end{pmatrix} = B^{-1}t^{2\times 1} = \frac{1}{C^2+(\theta\cdot D)^2}\begin{bmatrix}C & \theta\cdot D\\-\theta\cdot D & C\end{bmatrix}\begin{pmatrix}t_u\\t_w\end{pmatrix} \quad (14)$$

Next, $G$ of SIM(2) is a function of Lie algebra $\mathfrak{sim}(2)$:

$G(v) = exp\left(\sum_{i=1}^{4} v_i e_i\right)$. According to the $n$-th order derivative of the exponential function is continuous, so the group SIM(2) is continuous, differentiable, and smooth. The Lie group $G$ is continuous, differentiable, and smooth.

In addition, SO(2) and SE(2) are used for the comparison in rotation equivariance, the Lie group and Lie algebra forms of them are shown as follows,

$$SO(2) = \left\{R = \begin{bmatrix}cos\theta & -sin\theta\\sin\theta & cos\theta\end{bmatrix}\right\} \in \mathbb{R}^{2\times 2},$$

$\mathfrak{so}(2): \theta \in \mathbb{R}$

$$SE(2) = \left\{G = \begin{bmatrix}cos\theta & -sin\theta & t_u\\sin\theta & cos\theta & t_w\\0 & 0 & 1\end{bmatrix} = \begin{bmatrix}R^{2\times 2} & t^{2\times 1}\\0^{1\times 2} & 1\end{bmatrix}\right\} \in \mathbb{R}^{3\times 3}, \quad (15)$$

$\mathfrak{se}(2): \begin{pmatrix}u\\w\\\theta\end{pmatrix} \in \mathbb{R}^3$

### 4.2 Lie Group Convolutional Neural Network (Lie Group-CNN)

A Lie group-CNN is proposed to address the scale and rotation equivariance of image classification, as shown in Fig. 1. It consists of an initial lifting module, $N$ sequential group convolution modules, a global pooling layer, and a final classification layer. Details are presented as follows.

#### 4.2.1 Lifting module.

The lifting module consists of a lifting layer, a batch normalization (BN) layer, a nonlinear activation layer (ReLU is used here), a linear fully-connected layer, a batch normalization (BN) layer, and a nonlinear activation layer (ReLU) in sequence. The lifting layer transforms the input image from the $\mathbb{R}^n$ into the Lie group space, where the coordinates of the image are replaced by the Lie algebra of the Lie groups, e.g., SIM(2), SO(2), and SE(2), and the image intensity remains the same. The schematic of the lifting layer is shown in Fig. 2.

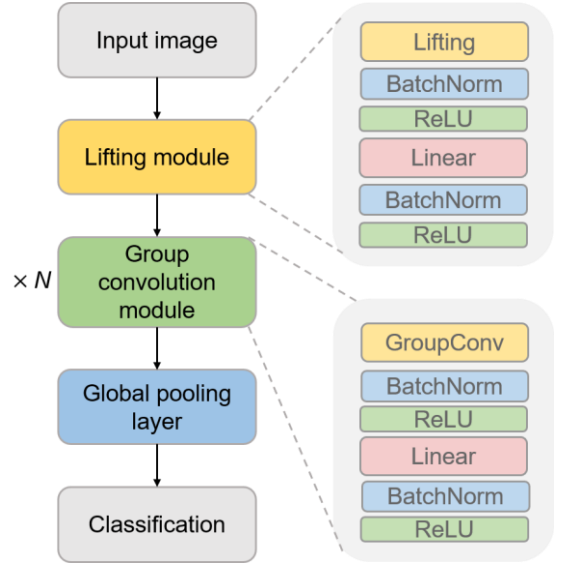

Fig. 1. The overall architecture of the proposed Lie group convolution neural network.

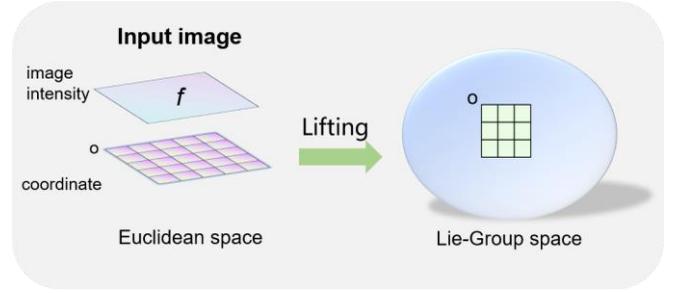

Fig. 2. Schematic of the lifting layer.

The lifting layer converts the image coordinates to group elements as

$$\{(x_i, f_i)\}_{i=1}^{N} \xrightarrow{Lifting} \{(g_{i,\kappa}, f_i)\}_{i=1,\kappa=1}^{N,K} \quad (16)$$

where $x$ denotes the coordinate in the $\mathbb{R}^n$ of the input image, $i$ denotes the index of coordinate point, $N$ denotes the total number of coordinate points ($N = H \times W$ for the input image), $f$ denotes the image intensity (for the gray-scale image, it is a scalar; for the RGB image, it denotes a three-dimensional vector), $g$ denotes the group element, and $K$ denotes the total number of group elements for each coordinate point. Since $g_1 g_2 \rightarrow g_3 \in G$, the lifting operation: $Lift(x_i) = \{g_{i,1},\ldots,g_{i,\kappa},\ldots,g_{i,K}\}$. It also shows that the image intensity does not change before and after the lifting operator, i.e., $f(x) = f(g)$.

Fixing any coordinate point as the origin $o$ in the $\mathbb{R}^n$, $o \in X$, if there exists a mapping function $\delta: G \rightarrow X$ and $\delta(g) = go$ is a surjection on $X$, then $X$ is called the homogeneous space of the group $G$, and an arbitrary point in the $\mathbb{R}^n$ can be obtained by transforming the origin $o$ using group elements in $G$. It forms the foundation of the lifting layer that can transform the input image from the $\mathbb{R}^n$ into the Lie group space.

All the group elements that can transform the origin $o$ to an arbitrary point $z$ are defined by the lifting operation: $\text{Lift}(z) = \{g \in G : go = z\}$. The group elements mapping the origin $o$ to itself are defined as the isotropy subgroup of $G$: $H = \{h \in G : ho = o\}$. For the left coset $gH$ of the isotropy subgroup $H$, $gho = go$ for any $h \in H$; therefore, the left coset $gH$ forms a homogeneous space $X$. Since the group space is closed, $\text{Lift}(z)$ often contains various group elements, and it is challenge to enumerate all possible group elements. A possible way to obtain rest group elements using a known group element $g_1$ is utilizing the isotropy subgroup to generate the left coset as $\text{Lift}(z) = \{g_1 H\}$.

### 4.2.2 Group convolution module.

After the lifting module, the original image has been mapped into Lie Group space. The next group convolution is operated on the Lie Group space. The group convolution module consists of a novel group convolution layer, a batch normalization (BN) layer, a nonlinear activation layer (ReLU is used here), a linear fully-connected layer, a batch normalization (BN) layer, and a nonlinear activation layer (ReLU) in sequence.

As shown in (9), the convolution kernel should be scaled and rotated in the opposite direction to obtain the scale and rotation equivariance, resulting in scale and rotation followed by a convolution differs from a convolution followed by scale and rotation. A scale and rotation equivariance group convolution is proposed to address this issue, which will be explained below.

Assume that the following convolution kernels exist in the group space after lifting operation of (16)

$$\psi_{A_{m_1}}(g) = \psi\left(A_{m_1}^{-1} g\right) \quad (17)$$

Substituting (17) into (6), we have

$$T_{m^{-1}} \psi_{A_{m_1}}(g) = \psi_{A_{m_1}}(A_m g) = \psi\left(A_{m_1}^{-1} A_m g\right) = \psi_{A_{m_1} A_m^{-1}}(g) \quad (18)$$

According to the multiplication closure of the group, $A_{m_1} A_m^{-1}$ is still a group element, showing that the convolution kernel is still located in the group space after scale and rotation transformations.

Substituting (18) into (9) yields

$$\left[(T_m f) * \psi_{A_{m_1}}\right](g) = T_m \left[f * \left(T_{m^{-1}} \psi_{A_{m_1}}\right)\right](g) = T_m \left[f * \psi_{A_{m_1} A_m^{-1}}\right](g) \quad (19)$$

Therefore, a scale and rotation equivariance group convolution is defined by (19) because the group convolution after scale and rotation equals the scale and rotation of a group convolution. It indicates that the convolution defined in the Lie group space can possess the scale, rotation and translation equivariance compared with the regular convolution in the $\mathbb{R}^n$. The (19) is specifically written in mathematical formulation of group convolution

$$(f * \phi)_G (q) = \int \phi\left(g^{-1} q\right) f(g) d\mu(g) \quad (20)$$

where $(f * \phi)_G$ denotes the group convolution of the input $f$ by a group convolution filter $\phi$, $g$ and $q$ denote the group elements of $G$, $g^{-1} q \in G$ according to the multiplication closure of the group, $f(g)$ denotes the output after the lifting operation, and $\mu$ denotes the Haar measure in the group space.

### 4.2.3 Parameterization of group convolution.

For the discrete group, all group elements can be enumerated and involved in the group convolution operation based on (20). However, it is challenge for the continuous Lie group. To fix this issue, a fully connected neural network is employed to parameterize the group convolution kernel as an approximated continuous function. Note that elements in Lie groups are in matrix forms and are not convenient to perform the tensor operation, while the corresponding Lie algebras are vector-wise and can interconvert with the Lie group elements by the exponential and logarithmic transformations. Inspired by this idea, the identity transformation $g = exp[log(g)] \in G$ is substituted into (20) as

$$\phi\left(g^{-1} q\right) = \phi \circ \left\{exp\left[log\left(g^{-1} q\right)\right]\right\} = (\phi \circ exp)\left[log\left(g^{-1} q\right)\right] \quad (21)$$

where $log(g^{-1} q)$ represents the Lie algebra of the group element $g^{-1} q$, and $(\phi \circ exp)$ is parameterized by a fully connected neural network noted as $\phi_\alpha : \mathfrak{g} \to \mathbb{R}$.

To determine the local region of the group convolution, the distance measure between two Lie group elements $g_1$ and $g_2$ is defined as

$$D(g_1, g_2) = \left\| log\left(g_1^{-1} g_2\right) \right\|_F \quad (22)$$

where $\|\cdot\|_F$ denotes the Frobenius norm.

Following the concept of reception field in the regular convolution, the group convolution operation is also defined within a local range of the continuous Lie group space as

$$\phi\left(g^{-1} q\right) = \begin{cases} \phi\left(g^{-1} q\right) & D(g,q) \leq r \\ 0 & D(g,q) > r \end{cases},$$

$$(f * \phi)_G (q) = \int_0^r \phi\left(g^{-1} q\right) f(g) d\left[D(g,q)\right] \quad (23)$$

Since $D(g,q) \leq r$ defines a local region in the Lie group space, (23) can be approximated by choosing a set of neighborhood group elements of $q$ to discretize the group convolution as

$$(f * \phi)_G (q) = \frac{1}{num\left[N(q)\right]} \sum_{g_i \in N(q)} \phi\left(g_i^{-1} q\right) f(g_i) \quad (24)$$

where $N(q)$ denotes the set of neighborhood group elements around $q$, $g_i$ denotes the $i$-th group element in $N(q)$, $num(\cdot)$ denotes the pre-set number of included group elements in the set.

The above distance measure remains invariant for the left multiplication of another group element, indicating the invariance of group convolution in the group space:

$$D(g_3 g_1, g_3 g_2) = log\left(g_1^{-1} g_3^{-1} g_3 g_2\right)_F = D(g_1, g_2) \quad (25)$$

For example, the distance measures between two elements in the SIM(2) group is

$$D(g_1, g_2) = \left\| log\left(g_1^{-1} g_2\right) \right\|_F \quad (26)$$

$$= \left\| \log \left[ \begin{pmatrix} s_1 \cdot cos\theta_1 & -s_1 \cdot sin\theta_1 & t_{u_1} \\ s_1 \cdot sin\theta_1 & s_1 \cdot cos\theta_1 & t_{w_1} \\ 0 & 0 & 1 \end{pmatrix}^{-1} \begin{pmatrix} s_2 \cdot cos\theta_2 & -s_2 \cdot sin\theta_2 & t_{u_2} \\ s_2 \cdot sin\theta_2 & s_2 \cdot cos\theta_2 & t_{w_2} \\ 0 & 0 & 1 \end{pmatrix} \right] \right\|_F$$

$$= \left\| \log \begin{pmatrix} s_2 \cdot cos\theta_2 & -s_2 \cdot sin\theta_2 & t_{u_2} \\ s_2 \cdot sin\theta_2 & s_2 \cdot cos\theta_2 & t_{w_2} \\ 0 & 0 & 1 \end{pmatrix} - \log \begin{pmatrix} s_1 \cdot cos\theta_1 & -s_1 \cdot sin\theta_1 & t_{u_1} \\ s_1 \cdot sin\theta_1 & s_1 \cdot cos\theta_1 & t_{w_1} \\ 0 & 0 & 1 \end{pmatrix} \right\|_F$$

$$= \left\| \begin{pmatrix} u_2 - u_1 \\ w_2 - w_1 \\ \theta_2 - \theta_1 \\ \lambda_2 - \lambda_1 \end{pmatrix} \right\|_F$$

$$= \sqrt{(u_2 - u_1)^2 + (w_2 - w_1)^2 + (\theta_2 - \theta_1)^2 + (\lambda_2 - \lambda_1)^2}, g_1, g_2 \in \text{SIM}(2)$$

In summary, the overall schematic of the proposed group convolution operation is shown in Fig. 3. 2× downsampling is performed to the Lie algebra and image intensity, and the farthest point sampling algorithm [37] is developed to facilitate the 2× downsampling process and shown in Algorithm 1. The dimension of the Lie algebra array reduces to $(bs, NK, NK, l)$, and the dimension of the image intensity array reduces to $(bs, NK, C)$, where $bs$ denotes the batch size, $N \times K$ denotes the total number of Lie algebra, $l$ denotes the length of the Lie algebra vector ($l = 1$ for the SO(2) group, $l = 3$ for the SE(2) group, and $l = 4$ for the SIM(2) group), $C$ denotes the channel number of the image intensity ($C = 3$ for the RGB image and $C = 1$ for the gray image after the lifting layer). The number of Lie algebra for each group element is set as $K = 1$ for simplification.

## 5 RESULTS AND DISCUSSION

### 5.1 Scaled-Rotated Dataset

In this section, two scaled-rotated datasets are utilized to verify the scale and rotation equivariance of the Lie group-CNN on SIM(2). Medical images have natural rotation and scale transformations due to the morphology similarity of targets with varying sizes and the lack of angular constraints between the observation device and objects. Therefore, two typical datasets of blood cells [38] and pigmentation skin diseases [39] are used to verify the scale and rotation equivariance of the proposed method.

*5.1.1 Blood cell dataset.*
The blood cell dataset consists of individual blood cell images acquired by the automatic analyzer and labeled by clinical pathologists. The dataset includes 11,959 training images, 1,712 validation images, and 3,421 test images and is labeled into eight categories of neutrophils, eosinophils, basophils, lymphocytes, monocytes, immature granulocytes, erythroblasts, and platelets. Each image is resized from 360×363×3 to 28×28×3. Some representative images are shown in Fig. 4.

An input image is converted into an array $\{x_i, f_i\}_{i=1}^{784}$ in the $\mathbb{R}^n$ and then fed into the Lie Group-CNN. The stochastic gradient descent algorithm is used for optimization, and the cross-entropy loss is used as the objective function to train a classifier on the dataset. The Lie group-CNN on SIM(2) that contains both rotation and scale group elements is utilized to verify the rotation-scale-equivariant capacity. In addition, the dilated convolution enabling multi-scale contextual information aggregation [23], STN [27], SESN using steerable filters to build scale equivariance convolutional networks [35], and L-Conv [36] are used for comparison. The training hyperparameters for the abovementioned networks remain the same and are determined by trial and error with an initial learning rate of 0.05, a batch size of 50, and a total epoch of 100.

Table 1 shows the comparison of the Top-1 test accuracy on the blood cell dataset. The results show that the accuracy of all methods is above 90%, and the Lie group-CNN on SIM(2) achieves the best recognition accuracy of 97.50% with 3.8%, 4.37%, 1.83%, and 7.32% improvement compared to the dilated convolution, STN, SESN, and L-Conv, respectively. The results indicate that the Lie group-CNN on SIM(2) can perform equivariant recognition on the blood cells with rotation and scale transformations, indicating that the proposed Lie group-CNN can better generalize and extract geometric features than Lie algebra, dilation convolution, spatial transformer, and steerable filter-based networks on rotated and scaled images.

*5.1.2 HAM10000 dataset.*
The HAM10000 dataset is a collection of dermatoscopic images of pigmented skin lesions obtained by the digital dermatoscopy system taken at different magnifications and angles. The dataset consists of 10,015 dermatoscopic images labeled into seven different categories and is split into 7,007 training images, 1,003 validation images, and 2,005 test images. Each image is resized from 600×450×3 to 28×28×3. The HAM10000 dataset has more shape variations than the blood cell dataset. Some representative images are shown in Fig. 5.

Similar to section 5.1.1, comparative studies are performed among the Lie group-CNN on SIM(2), dilation convolution, STN, SESN, and L-Conv and the training setup for the dataset is the same. Table 2 shows the comparison of the Top-1 classification accuracy on the HAM10000 dataset. The results show that the accuracy of all methods is in the range of 70-80% for the HAM10000 dataset. The results show that the accuracy of all methods is in the range of 70-80% for the HAM10000 dataset with more shape variations. However, the Lie group-CNN on SIM(2) achieves the best recognition accuracy of 77.90% with 3%, 5.85%, 2.1%, and 4.38% improvement compared to the dilated convolution, STN, SESN, and L-Conv, respectively. The results indicate that the Lie group-CNN on SIM(2) can also extract scale and rotation equivariance features from more complicated images on a relatively small dataset.

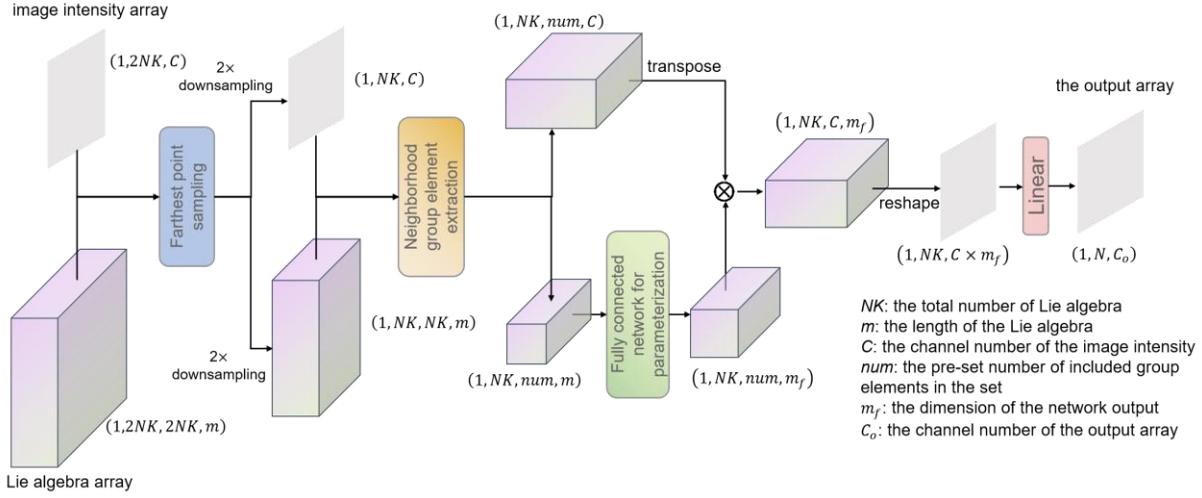

(a) Layouts of Lie group convolution

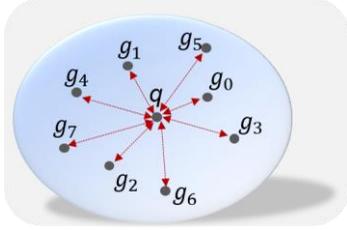

(b) Neighborhood group element extraction

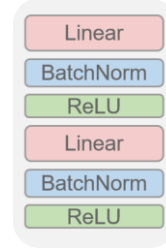

(c) Fully connected network for parameterization

Fig. 3. Schematic of the proposed group convolution operation.

---

**Algorithm 1** Farthest Point Sampling Algorithm

**Inputs**: A set of group elements $S = \{q_i\}_{i=1}^{num(S)}$, the number of required elements in $S_{sub}$ as $P = HW/2$.

Step 1: Choose an arbitrary group element from $S$ as the initial point $p_0$, $S_{sub} = \{p_0\}$.

Step 2: Calculate the distances between all the other group elements in $S$ and $p_0$.

Step 3: Form a distance set $D_0 = \{D_0^n\}, n = 1,\ldots,num(S)-1$.

Step 4: Select the farthest point with the maximum $D_0^n$ as $p_1$, and update the set $S_{sub} = \{p_0, p_1\}$.

Step 5: **for** $j=1, \ldots, P-1$ do

  Calculate the distance $D_j^n = D(q_n, p_j)$ for all $q_n \in S \setminus S_{sub}$

  Form a distance set $D_j = \{D_j^n\}, n = 1,\ldots,num(S \setminus S_{sub})$.

  if $D_j^n < D_0^n$

    $D_0^n = D_j^n$, select the farthest point with the maximum $D_0^n$ as $p_{j+1}$.

    Update the set $S_{sub} = \{p_0, p_1, \ldots, p_{j+1}\}$

**end**

**Return** $S_{sub}$

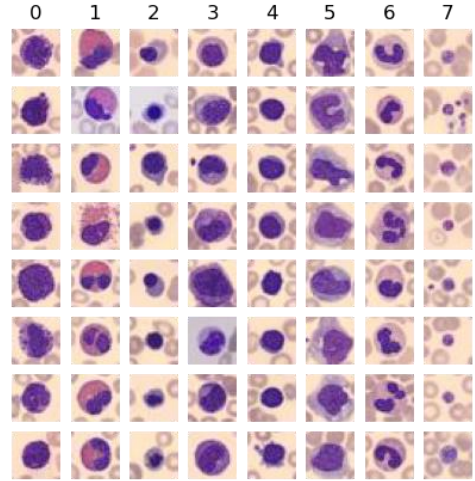

Fig. 4. Some representative images in the blood cell dataset.

### 5.2 Rotated Dataset

In this section, the rotated MNIST and the rotated CIFAR-10 datasets are utilized to verify the generalization ability of the method on rotation equivariance and the robustness of the network on the other two Lie groups.

TABLE 1

COMPARISON OF TOP-1 CLASSIFICATION ACCURACY ON THE BLOOD CELL DATASET

| Network | Top-1 accuracy on the blood cell dataset |
|---|---|
| Lie Group-CNN on SIM(2) | **97.50%** |
| Dilated Convolution [23] | 93.70% |
| STN [27] | 93.13% |
| SESN [35] | 95.67% |
| L-Conv [36] | 90.18% |

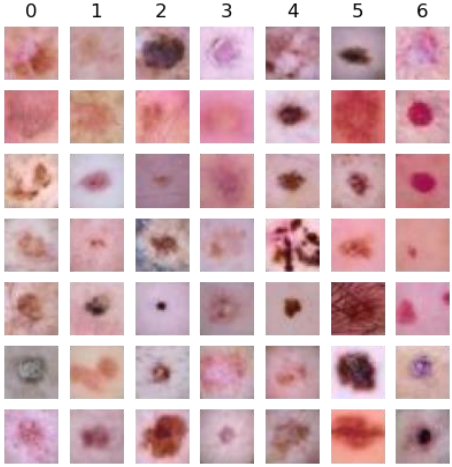

Fig. 5. Some representative images in the HAM10000 dataset

TABLE 2

COMPARISON OF TOP-1 CLASSIFICATION ACCURACY ON THE HAM10000 DATASET

| Network | Top-1 accuracy on the HAM10000 dataset |
|---|---|
| Lie Group-CNN on SIM(2) | **77.90%** |
| Dilated Convolution [23] | 74.90% |
| STN [27] | 72.05% |
| SESN [35] | 75.80% |
| L-Conv [36] | 73.52% |

### 5.2.1 Rotated MNIST dataset.

The rotated MNIST dataset [40] is generated by randomly rotating images in the MNIST dataset and used to verify the efficacy of the proposed rotation-equivariance Lie Group-CNN. It is split into a training set and a test set with 12,000 and 50,000 images, respectively. The image resolution is 28×28×1, and Fig. 6 shows some representative images.

A CNN classification model based on ResNet18 [3] is trained as the baseline. Three types of Lie groups, including SO(2), SE(2), and SIM(2) are investigated to demonstrate the effectiveness and generalization ability of the proposed method and the capacity of rotation equivariance. In addition, STN, L-Conv, and MLP-Mixer [41] are also compared to validate the superiority of the proposed method on the spatial transformer, Lie algebra, and MLP-based neural networks.

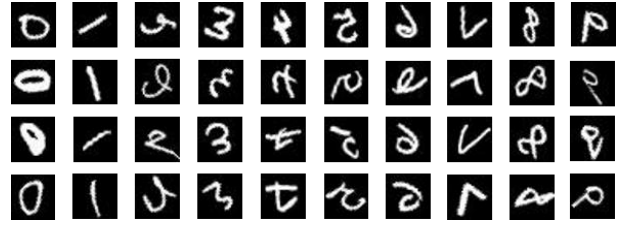

Fig. 6. Some representative images of the rotated MNIST dataset.

A CNN classification model based on ResNet18 [3] is trained as the baseline. Three types of Lie groups, including SO(2), SE(2), and SIM(2) are investigated to demonstrate the effectiveness and generalization ability of the proposed method and the capacity of rotation equivariance. In addition, STN, L-Conv, and MLP-Mixer [41] are also compared to validate the superiority of the proposed method on the spatial transformer, Lie algebra, and MLP-based neural networks.

Table 3 shows the comparison of the Top-1 test accuracy on the rotated MNIST dataset. The results show that the proposed Lie group-CNN on SO(2), SE(2), and SIM(2) outperforms the conventional ResNet18, STN, L-Conv, and MLP-Mixer, demonstrating the efficacy of rotation equivariance using the Lie group-CNN over the conventional CNN, spatial transformer, Lie algebra, and MLP-based neural networks. The generalization ability of the proposed method is also verified by the fact that using different Lie groups can improve classification accuracy against various rotation transformations. Among them, the Lie group-CNN on SO(2) achieves the best recognition accuracy of 98.2% on the rotated MNIST dataset, while using SE(2) and SIM(2) have slight decreases in the Top-1 accuracy. The results further indicate that the proposed Lie group convolution can better extract the embedding knowledge of symmetry than ResNet18, STN, L-Conv, and MLP-Mixer on rotated images.

TABLE 3

COMPARISON OF TOP-1 CLASSIFICATION ACCURACY ON THE ROTATED MNIST DATASET

| Network | Top-1 accuracy on the rotated MNIST dataset |
|---|---|
| Lie Group-CNN on SO(2) | **98.2%** |
| Lie Group-CNN on SE(2) | **96.8%** |
| Lie Group-CNN on SIM(2) | **97.4%** |
| ResNet18 [3] | 95% |
| STN [27] | 94.23% |
| L-Conv [36] | 91.79% |
| MLP-Mixer [41] | 83.9% |

### 5.2.2 Rotated CIFAR-10 dataset.

The original CIFAR-10 dataset consists of 50,000 training images in 10 classes. In this section, the rotated CIFAR-10 dataset is used to verify the efficacy of the proposed rotation-equivariance Lie Group-CNN. The training set of the rotated CIFAR-10 is obtained by randomly rotating the training images in the original CIFAR-10 dataset from 0 to 360 degrees, and the test set randomly samples 10,000 training images and rotates with different aspects to investigate the rotation equivariance. The resolution of the original images is 32×32×3, and the rotated images are padded with zeros to maintain an identical resolution. Some representative images are shown in Fig. 7.

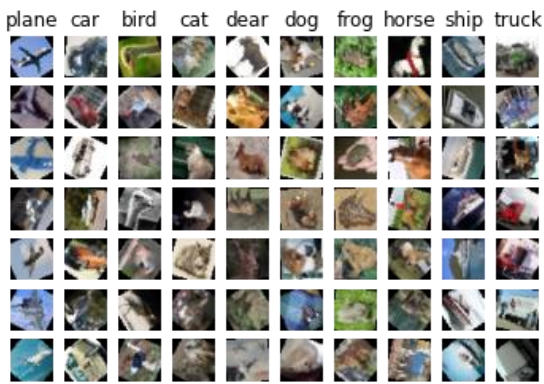

Fig. 7. Some representative images in the rotated CIFAR-10 dataset.

Similar to section 5.2.1, comparative studies are performed between three types of Lie groups, including SO(2), SE(2), and SIM(2), and the conventional ResNet18, STN, L-Conv, and MLP-Mixer. Table 4 compares the Top-1 classification accuracy on the rotated CIFAR-10 dataset using different classification networks. The Lie group-CNN on SO(2) achieves the best Top-1 classification accuracy on the rotated CIFAR-10 (99.9%), and the Lie group-CNN on SE(2) and SIM(2) only have a 0.1% decrease in classification accuracy which demonstrates the robustness of the Lie group-CNN on other Lie groups. The results show that using the proposed Lie group-CNN can significantly improve the classification accuracy on rotated images, further indicating the sensitivity of the Lie group convolution to the rotation equivariance. On the other side, model performances of ResNet18, STN, L-Conv, and MLP-Mixer on the rotated CIFAR-10 suggesting that the conventional CNN-based and MLP-based models have limited ability to extract symmetric knowledge on rotated images and fail to maintain the rotation equivariance.

TABLE 4

COMPARISON OF TOP-1 CLASSIFICATION ACCURACY ON THE ROTATED CIFAR-10 DATASET

| Network | Top-1 accuracy on the rotated CIFAR-10dataset |
| --- | --- |
| Lie Group-CNN on SO(2) | **99.9%** |
| Lie Group-CNN on SE(2) | **99.8%** |
| Lie Group-CNN on SIM(2) | **99.8%** |
| ResNet18 [3] | 79.7% |
| STN [27] | 62.83% |
| L-Conv [36] | 55.58% |
| MLP-Mixer [41] | 60.3% |

### 5.3 Visualization of Separability

To compare the separability of the aforementioned neural networks, the T-SNE algorithm [42] is utilized to reduce the high-dimensional feature maps from the last layer of each network to two dimensions for visualization. Figure 8 compares the separability visualization results on the blood cell dataset using the Lie group-CNN on SIM(2) and other different types of neural networks: L-Conv, dilated convolution, STN, and SESN. Eight categories of blood cells in various scales and rotations form the most distinct clusters in the Lie group-CNN on SIM(2), indicating that the proposed method performs the best on the blood cell dataset. The entangled impartibility of cluster boundary exists in other neural networks more or less with different severities, among which the L-Conv gains the most limited classification performance on the blood cell dataset. The results show that the proposed Lie group-CNN simultaneously obtains the best inter-class diversity and intra-class similarity and can extract invariant geometric features on images with scale and rotation transformations.

Figure 9 shows the visualization results of the proposed Lie Group-CNNs on the rotated datasets of MNIST and CIFAR-10. The results indicate that the Lie group-CNNs can effectively distinguish the different categories on these rotated datasets, among which the Lie group-CNN on SO(2) significantly performs the best than on SE(2) and SIM(2). The possible reason is that the extra learnable space caused by translation and scale elements in Lie algebra brings more challenges to rotation-only datasets.

## 6 RONCLUSIONS

This study proposes a novel Lie-group CNN to fix the issue of lacking scale and rotation equivariance for the regular CNN. A group convolution is designed in the Lie group space instead of the $\mathbb{R}^n$ to generalize scale and rotation equivariance. The main conclusions are obtained as follows:

(1) A Lie group-CNN with scale and rotation equivariance is constructed based on the group convolution, including an initial lifting module, $N$ sequential group convolution modules, a global pooling layer, and a final classification layer. The first lifting module transfers the input image from the $\mathbb{R}^n$ to the Lie group space. Theoretical analysis is performed to demonstrate that the group convolution after scale and rotation equals the scale and rotation of a group convolution, indicating that the proposed Lie-group CNN can possess the scale and rotation equivariance besides translation equivariance compared with the conventional CNN in the $\mathbb{R}^n$.

(2) The group convolution module consists of a novel group convolution layer, a batch normalization layer, a nonlinear activation layer, a linear fully-connected layer, a batch normalization layer, and a nonlinear activation layer in sequence. The Lie group elements can be transformed into the Lie algebras by surjective exponential mapping. For computational efficiency, the group convolution operators are locally defined using a distance measure between two Lie group elements, which remains invariant for the left multiplication of another group element. The group convolution operator is replaced by a fully connected network for parameterization in the Lie group space and facilitates the tensor operation.

(3) The SIM(2) group with both scale and rotation elements is utilized to construct a Lie group-CNN on SIM(2),

and the scale and rotation equivariance of the network is verified on two medical datasets, i.e., the blood cell dataset and the HAM10000 dataset. The results show that the Lie group-CNN on SIM(2) achieves the best recognition accuracy on both datasets compared to dilation convolution, spatial transformer, steerable filter, and Lie algebra-based networks. The results demonstrate that the Lie group-CNN on SIM(2) can perform equivariant recognition on objects with varying sizes and orientations and further indicate that the proposed method can extract geometric features on images with scale and rotation transformations.

(4) The robustness of the network is verified on other two Lie groups with rotation transformations (SO(2) and SE(2)), and the generalization ability of the Lie group-CNN on rotation equivariance is verified on two rotated image datasets. The results show that using the proposed Lie group-CNN on SO(2), SE(2), and SIM(2) can effectively improve the classification accuracy on rotated images, indicating the sensitivity of the Lie group convolution to the rotation equivariance and the robustness of the proposed method to different Lie groups.

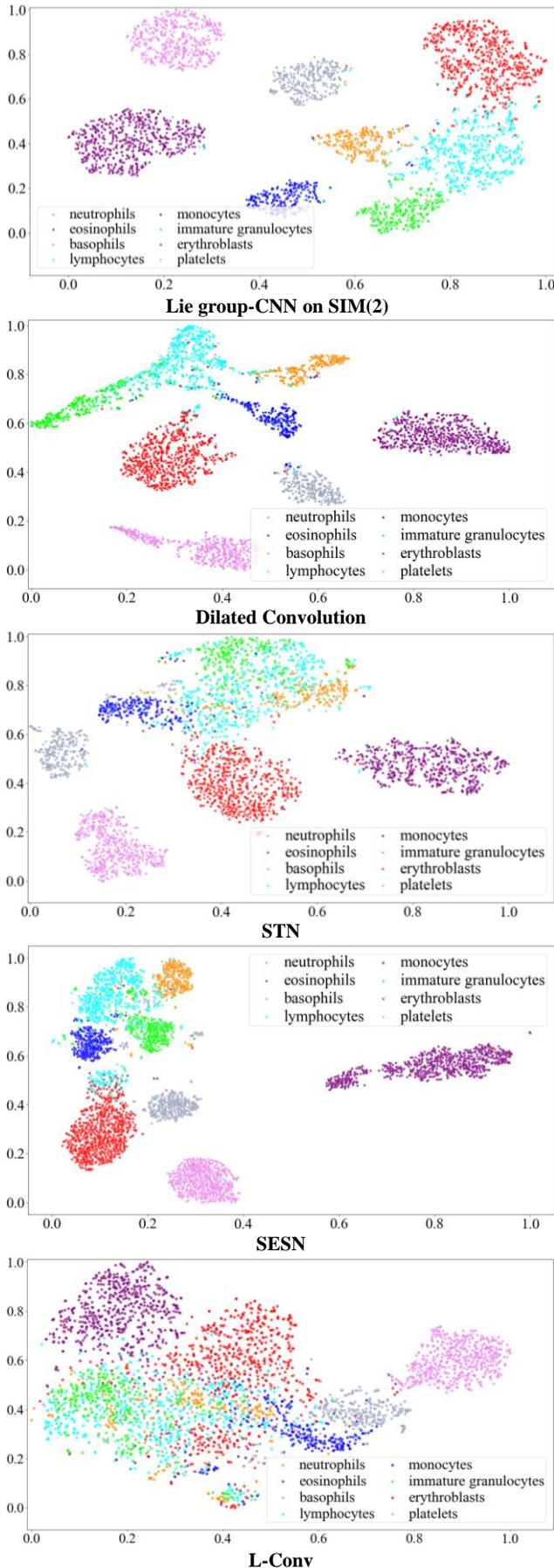

Fig. 8. Visualization of separability by multi-type neural networks on the blood cell dataset.

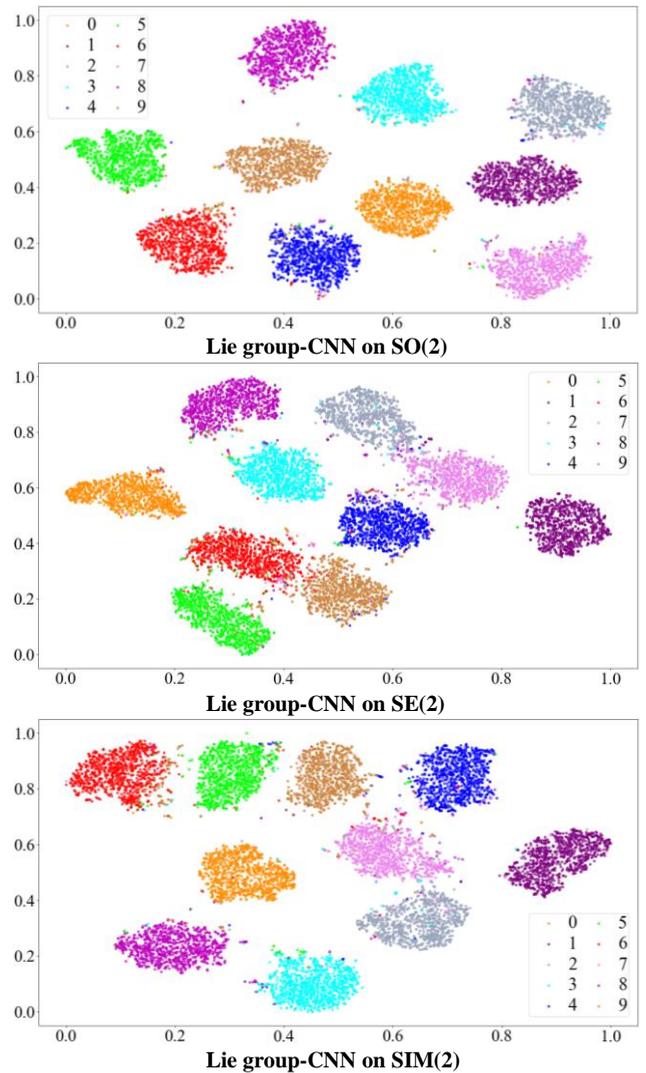

(a) the rotated MNIST dataset

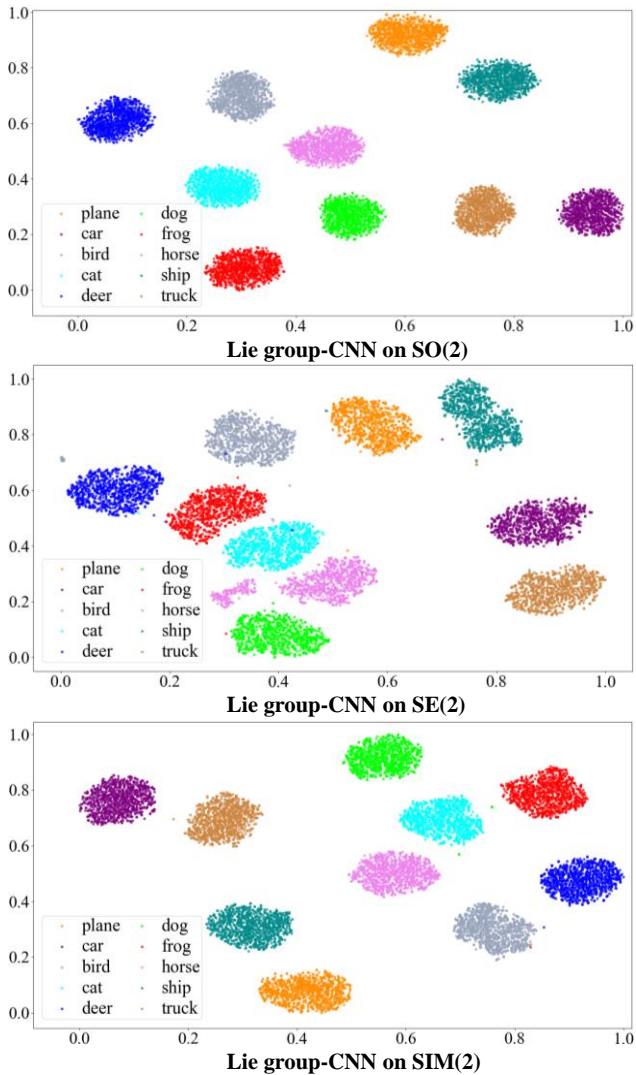

(b) the rotated CIFAR-10 dataset

Fig. 9. Visualization of separability of Lie group-CNNs on SO(2), SE(2), and SIM(2) on two rotated datasets.

The scale and rotation equivariance are enhanced by introducing the group convolution operator in the Lie group space instead of the conventional CNN in the $\mathbb{R}^n$, demonstrating the significance of group symmetry for feature extraction in image recognition tasks. In the future study, the ability to achieve more complex transformation equivariance will be further investigated and applied in object detection and image segmentation tasks.

## ACKNOWLEDGMENT

Financial support for this study was provided by the National Natural Science Foundation of China [Grant Nos. 51921006, 52192661, and 52008138], China Postdoctoral Science Foundation [Grant Nos. BX20190102 and 2019M661286], Heilongjiang Touyan Innovation Team Program, Heilongjiang Natural Science Foundation [Grant No. LH2022E070], and Heilongjiang Provincial Postdoctoral Science Foundation [Grant Nos. LBH-TZ2016 and LBH-Z19064].